\documentclass[german,english]{bvm2020}

\def\articlenumber{2774}

\date{}
\title{An Investigation of Feature-based Nonrigid Image Registration using Gaussian Process}

\titlerunning{Registration using Gaussian Process}

\author{Siming~Bayer$^{1,*}$, Ute~Spiske$^{1,*}$, Jie~Luo$^2$, Tobias~Geimer$^1$, William M.~Wells~III$^2$, Martin~Ostermeier$^3$, Rebecca~Fahrig$^3$,  Arya~Nabavi$^4$, Christoph~Bert$^5$, Ilker~Ey\"upoglo$^6$, Andreas~Maier$^1$}

\authorrunning{Bayer, Spiske et al.}

\institute{%
$^1$Pattern Recognition Lab, FAU Erlangen-N\"urnberg\\
$^2$Brigham and Women's Hospital, Harvard Medical School\\
$^3$Advanced Therapies, Siemens Healthare GmbH, Forchheim\\
$^4$Department of Neurosurgery, KRH Klinikum Nordstadt, Hannover\\
$^5$Department of Radiation Therapy, Universit\"at Klinikum Erlangen, Erlangen\\
$^6$Department of Neurosurgery, Univeris\"at Klinikum Erlangen, Erlangen\\
$^*$These authors contributed equally and are listed in alphabetical order}

\email{siming.bayer@fau.de}

\begin{document}

%
\selectlanguage{english}

\maketitle

\begin{abstract}
For a wide range of clinical applications, such as adaptive treatment planning or intraoperative image update, feature-based deformable registration (FDR) approaches are widely employed because of their simplicity and low computational complexity. FDR algorithms estimate a dense displacement field by interpolating a sparse field, which is given by the established correspondence between selected features. 
In this paper, we consider the deformation field as a Gaussian Process (GP), whereas the selected features are regarded as prior information on the valid deformations. Using GP, we are able to estimate the both dense displacement field and a corresponding uncertainty map at once. Furthermore, we evaluated the performance of different hyperparameter settings for squared exponential kernels with synthetic, phantom and clinical data respectively. The quantitative comparison shows, GP-based interpolation has performance on par with state-of-the-art B-spline interpolation. The greatest clinical benefit of GP-based interpolation is that it gives a reliable estimate of the mathematical uncertainty of the calculated dense displacement map.   
\end{abstract}

\section{Introduction}
For many clinical applications, nonrigid image registration is a key enabling technique. Compared to intensity-based deformable registration approaches, feature-based methods are intuitive and have low computational cost. Furthermore, human interactions, such as manual adjustment of landmarks, can be integrated easily. Basically, feature-based deformable registration methods estimate the suitable dense deformation field between two images by interpolating the correspondence of sparse feature sets.
An explicit interpolation step is therefore necessary to propagate the information from the control points to the whole image space. Hence, the choice of the interpolation technique and its underlying deformation model affects the overall performance of the registration method greatly.

One of the state-of-the-art interpolation technique is B-Spline interpolation~\cite{2774-01}.
It has been widely employed for various clinical applications such as registration of breast MR images~\cite{2774-01} or intraoperative brain shift compensation~\cite{2774-02}. 
Thin-Plate-Splines (TPS) proposed in~\cite{2774-03} are another common choice for deformable image warping, 
and has been applied e.~g.~for 
3D-3D~\cite{2774-04} registration of vasculature. 

Gaussian Process (GP) introduced in~\cite{2774-05} is a powerful tool to resolve regression, classification, and interpolation problems. The major advantage of Gaussian Process is the capability to estimate the result and its own uncertainty at once. Therefore, it has been applied in a wide variety of disciplines.
However, in medical image processing its applications to date are limited. Recently, \cite{2774-06} presented a generative model for intensity-based rigid registration with Gaussian processes, dealing with the interpolation uncertainty of the resampling step. A feature-based semi-automatic registration framework using Gaussian Process interpolation for the estimation of the dense displacement field has been proposed in~\cite{2774-07}. Although those works indicate the general applicability of GP for medical image registration, a comparison between GP and state-of-the-art image interpolation techniques and a comprehensive performance analysis is still missing.

We present a first investigation of the performance of GP for feature-based deformable registration, including comparison with B-Spline interpolation and analysis of hyperparameter setting. To this end, correspondence of selected features are established using the method proposed in~\cite{2774-02} and~\cite{2774-08}. Subsequently, GP with squared exponential kernel is employed to interpolate the sparse deformation field to a dense one and to estimate an associated uncertainty map. Furthermore, we compare three different approaches for the hyperparameter tuning. Finally, experiments with synthetic, phantom and clinical data for two clinical applications, namely intraoperative brain shift compensation and adaptive treatment planning for multi-catheter brachytherapy \cite{2774-08} are conducted for the performance analysis.

\section{Materials and Method}
The goal of image registration is to find a optimal transformation $\mathcal{T}$ maps the source image $\mathbf{I}_s$ to the target image $\mathbf{I}_t$. For the sake of simplicity, we only consider the case where $\mathbf{I}_s$ and $\mathbf{I}_t$ have the same dimensionality $\mathbb{R}^d$. The transformation $\mathcal{T}_{deformable}$ that maps $\mathbf{I}_s$ and $\mathbf{I}_t$ in a nonrigid fashion is a dense displacement field $\mathbf{V} = \lbrace \mathbf{v}_i\in\mathbb{R}^d, i=1:N \rbrace$ with $N$ denotes the size of $\mathbf{I}_s$. 

\subsection{Feature Extraction and Feature Matching}
Prior to the estimation of the dense deformation field, two sets of sparse features $\mathbf{P}=\lbrace \mathbf{x}_{i}^s\in \mathbb{R}^d,i=1:N_s \rbrace$ and $\mathbf{Q}=\lbrace \mathbf{y}_{i}\in\mathbb{R}^d,i=1:N_t \rbrace$ with $N_s$ and $N_t$ features are selected from $\mathbf{I}_s$ and the target $\mathbf{I}_t$, respectively. In the feature matching step, the source feature set $\mathbf{P}$ is updated to a corresponding set $\Tilde{\mathbf{P}} = \lbrace \Tilde{\mathbf{x}_{i}}\in \mathbb{R}^d,i=1:N_s \rbrace$, which is aligned with $\mathbf{Q}$. Consequently, a sparse displacement field maps the sparse feature sets $\mathbf{P}$ to $\mathbf{Q}$ is straightforwardly obtained as $\mathbf{D}=\lbrace \mathbf{d}_i = \Tilde{ \mathbf{x}_i} - \mathbf{x}_i,i=1:N_s\rbrace$.

\subsection{Estimation of Dense Deformation Field and Uncertainty Map} 
In order to interpolate $\mathbf{V}$ from $\mathbf{D}$, we consider the location of each voxel $\mathbf{x}_i$ in $\mathbf{V}$ as a multivariate Gaussian random variable. Each vector of the sparse displacement field $\mathbf{d}_i = \mathbf{d}(\mathbf{x}_i)$ for the voxel at the location $\mathbf{x}_i$ is treated as an observation of $\mathbf{V}$.  
According to the definition of GP provided in \cite{2774-05}, the prior distribution of the spatial position $\mathbf{x}_i$ is given as $\mathrm{d}(\mathbf{x}_i)\sim \mathrm{GP}(\mathrm{m}(\mathbf{x}_i),\mathrm{k}(\mathbf{x}_i,\mathbf{x}_j))$. Hereby, $\mathrm{m}(\mathbf{x}_i)=0$ is the mean function. The spatial correlation of the displacement vectors at the positions $\mathbf{x}_i$ and $\mathbf{x}_j$ is represented by the GP kernel $\mathrm{k}(\mathbf{x}_i,\mathbf{x}_j)$.

Following the GP modeling assumption, that the functions of all $\mathbf{x}_i$ in a set of random variables $\mathbf{X}$ are jointly Gaussian distributed, the dense displacement field can be formulated as a normal distribution 
$\mathrm{p}(\mathbf{V}\mid\mathbf{X})=\mathcal{N} (\mathbf{V}\mid\pmb\mu,\mathbf{K})$, with mean $\pmb\mu = (\mathrm{m}(\mathbf{x}_1) ,\dots,\mathrm{m}(\mathbf{x}_{N}))$, and covariance $\mathbf{K}=\lbrace K_{ij}=\mathrm{k}(\mathbf{x}_i,\mathbf{x}_j)\rbrace$. 
Consequently, the relationship between the $N_*=N-N_s$ unknown displacements $\mathbf{d_*}$ at the positions $\mathbf{X}_*$ and the $N_s$ known deformation vectors $\mathbf{d}$ at the locations $\mathbf{X}$ can be expressed as the  Equation.~\ref{gaussian}, where $\mathbf{K}=\mathrm{k}(\mathbf{X},\mathbf{X})$, $\mathbf{K}_*=\mathrm{k}(\mathbf{X},\mathbf{X_*})$, and $\mathbf{K_{**}}=\mathrm{k}(\mathbf{X_*} ,\mathbf{X_*})$ are covariance matrices with the size $N_s \times N_s$, $N_s \times N_*$ and $N_* \times N_*$ respectively. 
	\begin{equation}
	\label{gaussian}
	    \begin{pmatrix}
		\mathbf{d}\\
	    \mathbf{d}_* \\
	    \end{pmatrix}
	    \sim \mathcal{N}\left(
	    \mathbf{0},
	    \begin{pmatrix}
		\mathbf{K} &	  \mathbf{K}_*\\
		\mathbf{K}^T_* &	  \mathbf{K}_{**}\\
	    \end{pmatrix}
	    \right)
	\end{equation}
Having the observations $\mathbf{d}$, the prior GP assumption can be converted into GP posterior $p(\mathbf{d}_*\mid\mathbf{X_*},\mathbf{X},\mathbf{d})= \mathcal{N}(\mathbf{d}_*\mid\pmb\mu_*,\pmb\Sigma_*)$ via bayesian inference, where $\pmb{\mu}_*= \mathbf{K}^T_*\mathbf{K}^{-1}\mathbf{d}$. Simultaneously, an uncertainty map indicates the mathematical confidence of the estimated displacement vectors can be obtained from the diagonal entries of the covariance matrix $\pmb{\Sigma}_*$, defined in Equation.~\ref{cov}).

\begin{equation}
\label{cov}
\pmb\Sigma_*= \mathbf{K}_{**}-\mathbf{K}^T_*\mathbf{K}^{-1}\mathbf{K}_*.
\end{equation}





 
\emph{GP kernel estimation:}
The zero mean assumption of GP implies, that GP is completely defined by its second-order statistics \cite{2774-09}. 
Hence, the choice of the kernel and its parameter setting are the key factors defining the behavior of the GP model.
In order to preserve the smoothness of the resulting dense displacement field, we use a Squared Exponential kernel, $\mathrm{k}(x, x')=\sigma^2exp(-\frac{\lVert x-x'\rVert^2}{2l^2})$ (also known as Gaussian kernel). The characteristic of the Squared Exponential (SE) kernel is defined by $l$ and $\sigma$. The former is the length-scale of the random variable that controls the smoothness of the kernel, and the latter represents the relationship between output displacement vectors.
In this work, we use three different method for automatic calculation of the hyperparameters $l$ and $\sigma$:
\begin{description}
    \item [\texttt{MEAN}] Initially, the standard deviation of the kernel is computed as the mean standard deviation of the sparse displacement vectors used to train the GP model. It can be expressed as
    $\sigma_{mean} = \mathrm{mean}\{\sigma_{disp} :=\sigma(\mathbf{D}_x), \sigma(\mathbf{D}_y), \sigma(\mathbf{D}_z)\}$. The length-scale is initialized as $l_{mean} =\lbrace \mathrm{mean}(\lVert \mathbf{x}_i-\mathbf{x}_j\rVert^2),i\neq j\wedge i, j = 1:N_s, \mathbf{x}\in\mathbb{R}^d\rbrace$.
    \item [\texttt{NML}]Negative log marginal likelihood minimization proposed by \cite{2774-08}. The start parameters are estimated using the \texttt{MEAN} method.
    \item [\texttt{DGS}] Use discrete grid search to optimize the hyperparamter $\sigma$ and $l$ in the following discrete space:
    \begin{itemize}
        \item $\sigma_{dgs}\in\{\mathrm{min}(\sigma_{disp}),\sigma_{mean},\mathrm{max}(\sigma_{disp})\}$,
        \item $l_{dgs}\in\{\mathrm{min}(\lVert \mathbf{x}_i-\mathbf{x}_j\rVert^2), l_{mean}, \mathrm{max}(\lVert \mathbf{x}_i-\mathbf{x}_j\rVert^2),i\neq j\wedge i, j = 1:N_s\}$
    \end{itemize} 
    The objective function of the grid search optimization is the root mean squared error (RMSE) between the updated image $I_{warp}$ and the target image $I_{target}$.
\end{description}

\section{Results}

\begin{table}[t]
   \scriptsize
   \caption{Summary of the conducted quantitative experiments with synthetic, phantom, and clinical data. Clinical data has different size in the axial direction, but has the same resolution as the phantom data.} 
   \label{tab:metrics}
   \centering 
   \begin{tabular}{lcccccc} 
   \hline\hline
         \emph{Data} & \emph{Voxels}  & \emph{No.~Features}  & \emph{\% Used features}  & \emph{No.~Image pairs}&\emph{Metric}\\ 
   \hline
    Synthetic & $256^3$ & $[1000, 2000]$ & $20\%$ &6&$\frac{\text{No. }(I_{warp} \not\equiv I_{target})}{\text{No.Voxel}} $ \\
    Phantom & $512^3$ & $[3000, 4000]$ & $20\%$ &4& MHD\cite{2774-02} \\
    Clinical & $512\times512\times$n & $[200, 400]$ & $100\%$ &6&$\frac{|\Delta\text{HU}|}{\text{No.Voxel}}$\\
    
   \hline
   \end{tabular}
\end{table}
In order to evaluate the accuracy and applicability of GP for different medical applications, we conduct quantitative experiments with \textit{synthetic data} \cite{2774-10} and \textit{anthropomorphic phantom data} \cite{2774-11} for intraoperative brain shift compensation. For the adaptive treatment planning for multi-catheter HDR Brachytherapy, a retrospective clinical study is conducted. 

A summary of the conducted quantitative experiments is presented in Table.~\ref{tab:metrics}. The vessel centerline of the \textit{synthetic data} and \textit{phantom data} are extracted and aligned using the framework proposed in \cite{2774-02}. The catheters in the \textit{clinical data} are registered as described in \cite{2774-08}. Since the number of the homologous features differs in the three categories of experiments greatly, we randomly select $20\%$ features from the registered vessel centerlines from the synthetic and phantom data for a fair comparison. Considering the image properties \footnote{Synthetic data are binarized (background: $0$; brain parenchyma: $255$) \cite{2774-10}, phantom data have unrealistic Hounsfield Unit (HU) value \cite{2774-11}}, number of homologous features and the availability of the ground truth, we calculated modified hausdorff distance (MHD) between the extracted vessel centerlines from the warped and target image for the evaluation of phantom experiments. For the synthetic and clinical data, pro voxel intensity difference are estimated. 

\begin{figure}[!htbp]
    \centering
    \includegraphics[width=\textwidth]{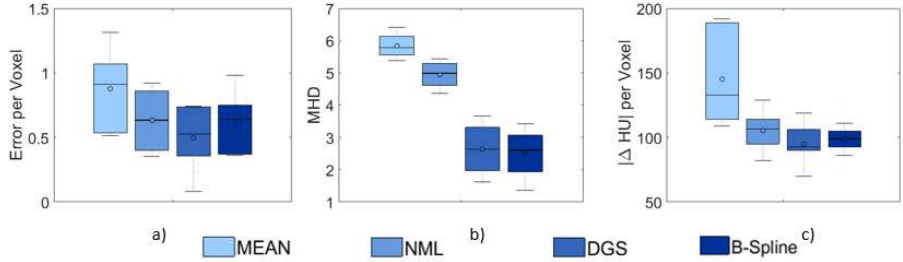}
    \caption{Quantitative results of all experiments conducted. \texttt{MEAN}, \texttt{NML} and \texttt{DGS} are compared with B-Spline interpolation. For phantom data b), MHD of the vessel centerline extracted from the warped and target image is calculated. Average intensity difference is used as metrics for synthetic a) and clinical c) data.}
    \label{fig:error}
\end{figure}
\begin{figure}[ht]
    \centering
    \includegraphics[width=\textwidth]{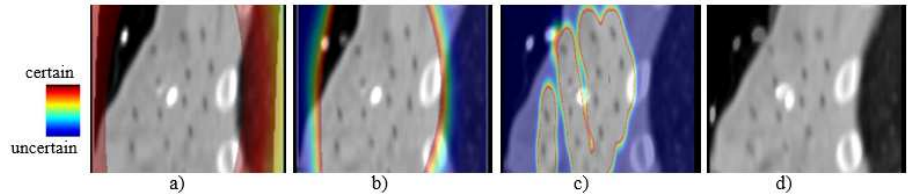}
    \caption{Overlay of the warped clinical images and their corresponding uncertainty map (color map). The dense displacement fields are interpolated from the sparse field (established with the corresponding catheter points, i.e. the black dots on the images) using a) \texttt{MEAN}, b), \texttt{DGS} c) \texttt{NML}, and d) B-Spline respectively. }
    \label{fig:brachy_visualization}
\end{figure}

The performance of \texttt{MEAN}, \texttt{NML} and \texttt{DGS} are compared with the state-of-the-art B-Spline interpolation technique \cite{2774-01}. The quantitative results are presented in Fig.~\ref{fig:error}. For the qualitative inspection, overlays of the warped clinical images and their corresponding uncertainty maps are presented in Fig.~\ref{fig:brachy_visualization}. 


\section{Discussion}
The quantitative results of all experiments conducted show the same trend, namely \texttt{DGS} and B-Spline interpolation outperforming \texttt{MEAN} and \texttt{NML}. Moreover, \texttt{DGS} outperforms B-Spline  slightly both for synthetic and clinical data. For phantom data, \texttt{DGS} and B-spline show comparable results. These results indicate, that modeling nonrigid deformation as a GP with Squared Exponential kernel presents a reliable alternative to B-Spline interpolation in terms of accuracy. 

Additionally, the qualitative results of the cinical data presented in Fig.~\ref{fig:brachy_visualization} underline our quantitative findings, namely, the result of \texttt{DGS} and B-Spline being comparable. The uncertainty map on Fig.~\ref{fig:brachy_visualization} a), b), and c) visualize the mathematical confidence of the estimated voxel-wise displacement. Hereby, \texttt{MEAN} and \texttt{NML} tend to be overconfident about its own estimation of dense displacement, whereas \texttt{DGS} produces a uncertainty map with more credability. Considering both quantitative and qualitative results, a conclusion about the self-confidence of GP interpolation with Squared Exponential kernel can be drawn: with the suitable choice of hyperparameter, GP-based interpolation is comparable with B-Spline interpolation, in term of accuracy. More importantly, it produce a confidence map about its own estimation with high credability, which could be used as a guidance for clinicians.

In general, GP belongs to the family of non-parametric methods. For the interpolation of each unknown displacement vector, the entire set of training data is taken into account. Hence, the run time of GP depends on the size of the sparse displacement vectors and the size of the image. In contrast, B-Spline interpolation is locally controlled, which means only the number of the homologous points on a predefined mesh grid affects its computational cost. Consequently, B-Spline interpolation is suitable for time critical applications such as intraoperative brain shift compensation, whereas GP-based interpolation is an excellent candidate for applications with lower real-time requirement, such as treatment planning in radiation therapy. In the subsequent studies, a detailed run time analysis of the GP-based interpolation technique will be performed. Furthermore, qualitative studies with clinicians are planned, where the clinician will ask to give scores for different interpolation methods. 






\bibliographystyle{bvm2020}

\bibliography{2774}
\end{document}